\documentclass[letterpaper, 10 pt, conference]{ieeeconf}  

\IEEEoverridecommandlockouts                              

\usepackage{booktabs}       
\usepackage{amsfonts}       
\usepackage{nicefrac}       
\usepackage{pifont}
\usepackage{xcolor}         
\usepackage{multirow}
\usepackage{microtype}      
\usepackage{graphicx} 
\usepackage{amsmath}
\usepackage{amssymb}
\usepackage{wrapfig}
\usepackage{colortbl}
\usepackage{url}
\usepackage[colorlinks,citecolor=ForestGreen]{hyperref}

\def\etal{\emph{et al}.}

\newcommand{\myParagraph}[1]{\noindent \textbf{#1} ---}

\definecolor{Gray}{gray}{0.85}
\definecolor{GrayBorder}{gray}{0.65}
\definecolor{green_cylinder}{rgb}{0.0,0.40,0.0}
\definecolor{blue_cylinder}{rgb}{0.02,0.05,0.75}
\definecolor{way_point}{rgb}{0.56,0.0,1.0}
\newcolumntype{a}{>{\columncolor{GrayBorder}}c}
\newcolumntype{b}{>{\columncolor{Gray}}r}

\title{\LARGE \bf
Multi-Object Navigation in real environments using hybrid policies
}

\author{Assem Sadek$^{1}$, Guillaume Bono$^{1}$, Boris Chidlovskii$^{1}$, Atilla Baskurt$^{2}$ and Christian Wolf$^{1}$
\thanks{
We recognize support through French grant ``\textit{Remember}'' (ANR-20-CHIA0018) of call ``\textit{Chaires IA hors centres}''.
}
\thanks{$^{1}$Naver Labs Europe, Meylan, France. 
        {\tt\small firstname.lastname@naverlabs.com}}%
\thanks{$^{2}$Université de Lyon, INSA-Lyon, LIRIS, Villeurbanne, France.
        {\tt\small atilla.baskurt@insa-lyon.fr}}%
}

\begin{document}

\maketitle
\thispagestyle{empty}
\pagestyle{empty}

\begin{abstract}
        Navigation has been classically solved in robotics through the combination of SLAM and planning. More recently, beyond waypoint planning, problems involving significant components of (visual) high-level reasoning have been explored in simulated environments, mostly addressed with large-scale machine learning, in particular RL, offline-RL or imitation learning. These methods require the agent to learn various skills like local planning, mapping objects and querying the learned spatial representations. In contrast to simpler tasks like waypoint planning (PointGoal), for these more complex tasks the current state-of-the-art models have been thoroughly evaluated in simulation but, to our best knowledge, not yet in real environments.

        In this work we focus on sim2real transfer. We target the challenging Multi-Object Navigation (Multi-ON) task~\cite{DBLP:conf/nips/WaniPJCS20} and port it to a physical environment containing real replicas of the originally virtual Multi-ON objects.
        We introduce a hybrid navigation method, which decomposes the problem into two different skills: (1) waypoint navigation is addressed with  classical SLAM combined with a symbolic planner, whereas (2) exploration, semantic mapping and goal retrieval are dealt with deep neural networks trained with a combination of supervised learning and RL. We show the advantages of this approach compared to end-to-end methods both in simulation and a real environment and outperform the SOTA for this task~\cite{MarzaIROS2022}.

\end{abstract}

\section{Introduction}
\label{sec:introduction}
\noindent
Robot navigation has progressed from simple waypoint navigation problems in richly prepared environments, for which robustly working systems are now used in production, to complex tasks 
involving high-level reasoning with visual and semantic concepts. This has been made possible through large-scale machine learning, mostly in photo-realistic simulators and reinforcement learning from billions of interactions. The resulting robotic agents, implemented as high-capacity neural networks, 
which are however subject to performance drop
and lack of robustness when the policies are transferred from simulation to real environments with physical robots. This is mostly due to the gap in realism between simulation and reality ("sim2real gap"), as well as the difficulty to explore a large amount of variation factors inherent in navigation problems, such as room layouts, furniture, textures and other room details, rare local scene geometries etc. 

There is a recent trend towards modular approaches which decompose the problem into hierarchical parts~\cite{Chaplot2020Learning,BeechingECCV2020} and hybrid approaches, which combine a   shortest path planner (symbolic or trained) with a trained policy. While these approaches have been shown to be more sample efficient~\cite{ramakrishnan_poni_2022}, state-of-the-art methods are still evaluated in simulation and lack thorough tests on real  robots.

\begin{figure}[t] \centering
\includegraphics[width=\linewidth]{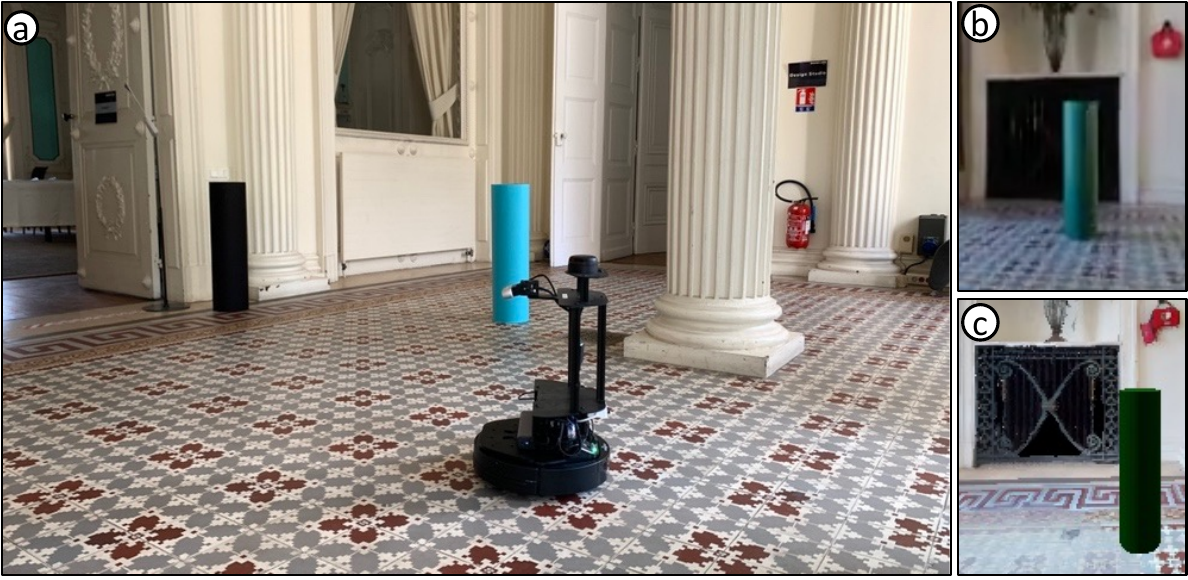}
\caption{\label{fig:chateau}We perform Multi-Object Navigation \cite{DBLP:conf/nips/WaniPJCS20}, i.e. the sequential visual search of multiple object in a given order, and are the first do this in real physical environments (a) characterized by a large sim2real gap. This is illustrated by two first-person views (b), real and (c), simulation. We propose a hybrid method combining classical mapping and deep learning, and compare to the SOTA methods on this task using end-to-end RL training and auxiliary losses \cite{MarzaIROS2022}.}
\end{figure}

\begin{figure*}[t]
\includegraphics[width=0.95\textwidth]{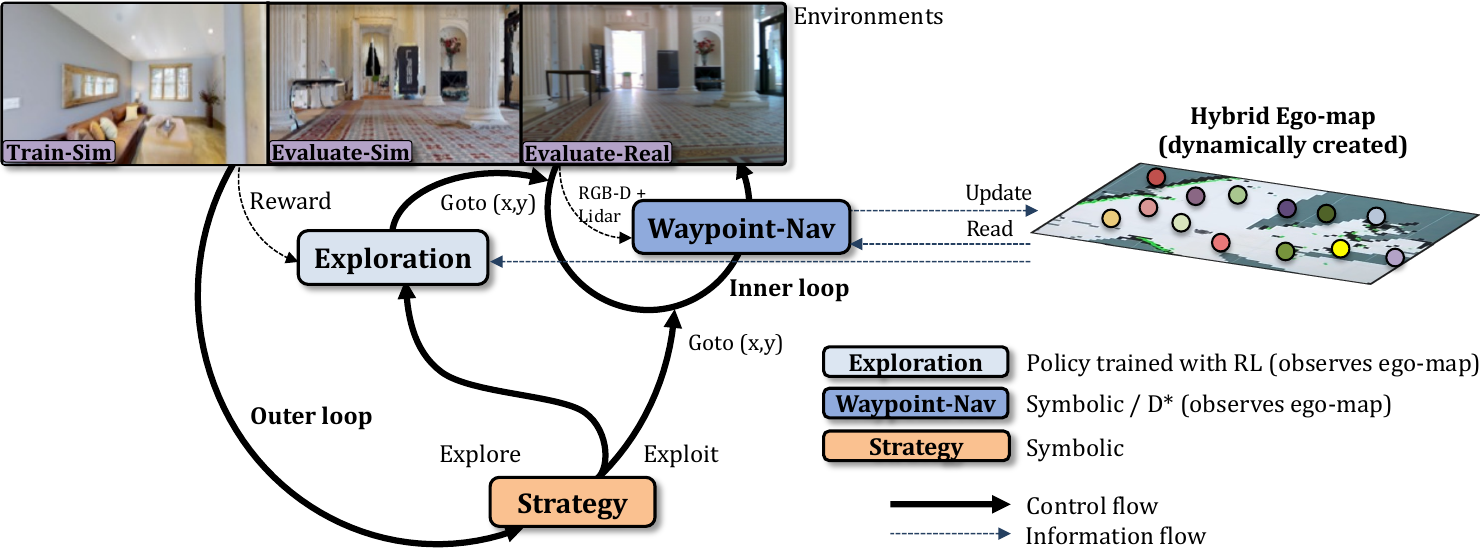}
\caption{\label{fig:teaser}An agent for multi-object navigation maintains a hybrid representation consisting of a metric bird's eye view map combined with a semantic point cloud. The agent switches between a trained exploration policy and symbolic waypoint selection, deferring low-level actions to a symbolic planner.}
\end{figure*}

In this work we address the challenging problem of Multi-Object Navigation~\cite{DBLP:conf/nips/WaniPJCS20}, which, similarly to the K-items scenario~\cite{Beeching2020ICPR}, requires an agent to sequentially navigate through a set of objects in an imposed order. This task definition favors agents capable of learning to map seen objects in an internal spatial representation, as navigating to them later in the episode can increase reward. This makes it stand out with respect to simpler  tasks like ObjectNav, where the combined capacities of exploration and reactive local planning from the current observation are sufficient to solve the task\footnote{Regularities in spatial layouts may be exploited with an additional form of higher reasoning, for instance with potential fields~\cite{ramakrishnan_poni_2022}, but we do not focus on these aspects in this work.}.

We target sim2real transfer and, to our best knowledge, are the first to perform a thorough performance evaluation of a method on Multi-Object Navigation in a real physical environment, see Figure~\ref{fig:chateau}. 
While simpler tasks, such as PointGoal, have been evaluated on real robots \cite{kadian_sim2real_2020,SadekICRA2022}, evaluation of trained models on more complex tasks has been sparse or nonexistent.
We present a new method for navigation, whose design choices have been driven by the objective of optimizing performance in real environments. We propose a new hybrid method which decomposes the problem into two  parts: 
\begin{itemize}  
\item[\ding{192}]\textbf{``Good Old Fashioned Robotics''(GOFR)}, that deals with classical navigation aspects not related to semantics, such as detection of navigable space and localization (geometric SLAM) combined with waypoint navigation on this map.
\item[\ding{193}]\textbf{Semantics through Machine Learning}, i.e. mapping semantic concepts required for visual reasoning and exploiting them; exploration of the most promising areas of the environment exploiting layout regularities.
\end{itemize}
During navigation, a classical SLAM algorithm~\cite{labbe19rtabmap} creates and maintains a 2D metric representation in the form of a tensor/map and localizes the robot on it using Lidar input. High-level features, extracted from visual RGB-D observations with a deep neural network, form a spatial and semantic point cloud, whose spatial coordinates are aligned with the metric representation, see Figure \ref{fig:teaser}. The combined hybrid representation satisfies the needs of relevant sub-skills the agent requires: (i) to determine whether a target object has been observed in the past, (ii) to plan optimal trajectories between the agent and explored areas, and (iii) to determine the frontiers of unexplored areas in the environment and thus the next intermediate sub-goals in case the environment needs to explore to find the next goal. All these sub-skills are designed and trained separately, which allows to limit sample complexity of training. 

The contributions of this work are the following:
(i) we introduce a hybrid method for Multi-Object navigation combining classical metric SLAM and path planning with learned components trained with supervised learning and RL; 
(ii) we reproduce the Multi-ON benchmark~\cite{DBLP:conf/nips/WaniPJCS20} in a real environment, where we place manufactured reproductions of the goal objects, used in originally simulated target environment;
(iii) we compare the proposed method to end-to-end trained methods in this real environment, in particular with the winning entry of the CVPR 2021 Multi-ON competition~\cite{MarzaIROS2022}, which we outperform in both real and simulated environments.

\section{Related work}
\label{sec:sota}

\myParagraph{Modular Embodied Navigation} 
Casting a task as an end-to-end learning problem is widely used in CV and NLP. Navigation has been addressed through this lens early on, e.g. for exploration~\cite{Chen19exploration}, where a neural policy processes raw sensory observations and directly predicts agent actions.
However, jointly learning mapping, state-estimation and path-planning purely from data has been shown to be expensive~\cite{Chaplot2020Learning}. An alternative are hierarchical and hybrid architectures~\cite{BeechingECCV2020,Chaplot2020Learning,Chaplot20objectgoal,Chaplot2020semantic} that compose a learned mapper with global and local policies, where all components interface via the map and an analytical path planner. Pushing modularity further, Ramakrishnan~\etal~\cite{ramakrishnan_poni_2022} propose to disentangle the skills of ‘\textit{where to look?}’ 
from  navigation itself. 
The network predicts potential functions conditioned on a semantic map and uses them to decide where to look for an unseen object. 

\myParagraph{Sim2Real} 
The sim2real gap can compromise strong performance achieved in simulation when agents are tested in the real-world~\cite{hofer2020sim2real}, as perception and control policies often do not generalize well to real robots due to  inaccuracies in modelling, simplifications and biases. To close the gap, domain randomization methods~\cite{peng18sim,tan18real} treat the discrepancy between the domains as variability in the simulation parameters. 
Alternatively, domain adaptation methods learn an invariant mapping for matching distributions between the simulator and the robot environment. Examples include transfer of visuo-motor policies 
by adversarial learning~\cite{zang19adversarial}, adapting dynamics in RL~\cite{eysenbach21off}, and adapting object recognition to new domains~\cite{zhu19adapting}. Bi-directional adaptation is proposed in~\cite{truong_bi-directional_2021}; Recently, Chattopadhyay~\etal~\cite{chattopadhyay2021robustnav} benchmarked the robustness of embodied agents to visual and dynamics corruptions.
Kadian~\etal~\cite{kadian_sim2real_2020} investigated sim2real predictability of Habitat-Sim~\cite{Savva_2019_ICCV} for PointGoal navigation and proposed a new metric to quantify it, called Sim-vs-Real Correlation Coefﬁcient (SRCC). The PointGoal task on real robots is also evaluated in~\cite{SadekICRA2022}. To reach competitive performance without modelling any sim2real transfer, the agent is pre-trained on a wide variety of environments and then fine-tuned on a simulated version of the target environment. 


\myParagraph{Memory and Maps (Inductive Bias)} 
Memory is a crucial aspect of an intelligent agent’s ability to reason about 3D space and geometry. Neural memories like NeuralMap~\cite{parisotto18neuralmap}, MapNet~\cite{henriques18mapnet} and propose latent metric maps,
which are updated incrementally from the camera observations and odometry and act as inductive bias for end-to-end training.
EgoMap~\cite{BeechingECCV2020} 
augments these maps with multi-step objectives and attention reads, trained with RL. Alternative neural maps are topological maps \cite{savinov2018semiparametric,Chaplot_2020_CVPR,BeechingECCV2020}, transformers \cite{EpisodicTransformersICCV2021,Fang_2019_CVPR,DecisionTransformer2021,OfflineTransformer2021,chen_think_2022,reed_generalist_2022}, which break the Markovian assumption and attend to a large temporal horizon and implicit representations \cite{ImplicitNavICLR2022,RobNavNERF2022}.

\myParagraph{Exploration} 
 is at the core of all navigation tasks \cite{Chaplot2020Learning} and is in itself studied and evaluated~\cite{anderson2018evaluation}.  Efficiently visiting the environment is useful for solving tasks in known environments and pre-mapping in unknown ones. 
Chen~\etal~\cite{Chen19exploration} explored policies with spatial memory that are bootstrapped with imitation learning and finetuned with coverage reward.
In \cite{Chaplot2020semantic}, an exploration policy is trained by introducing semantic curiosity based on observation consistency.
%
SEAL~\cite{Chaplot2021SEAL} trains perception models on internet images to learn an active exploration policy. They build 3D semantic maps to learn both action and perception models, and integrate intrinsic motivation.
Episodic semantic maps are proposed in~\cite{Chaplot20objectgoal}.

\section{Hybrid planning and navigation}
\label{sec:method}
\noindent
We target the task of Multi Object Navigation (Multi-ON) introduced by Wani~\etal~\cite{DBLP:conf/nips/WaniPJCS20}, in particular the 3 object variant: during each episode, the agent has to find 
3 cylindrical objects $G_n$, $n=1,2,3$, in a pre-defined order, where $G_n$ is the $n^{th}$ object to find,  and is required to call the \textit{Found} action at each goal. The episode duration is limited to 2,500 environment steps. At each step $t$, the agent receives an egocentric RGB-D observation $\mathbf{O}_{t} \in \mathbb{R}^{h \times w \times 4}$, a Lidar frame, and the class label of the current target object taken from 8 classes. 
All training was performed in simulation only with the Habitat simulator~\cite{Savva_2019_ICCV}, but the system was evaluated, both, in simulation and on a real Locobot robot in a real environment, more details are given in Section~\ref{sec:experiments}.

With operations on robots in real environment and conditions in mind, we follow a modular approach, outlined in Figure~\ref{fig:teaser}. The method is hybrid; it  leverages both trained neural modules for perception and exploration, and classical algorithms for occupancy mapping, localization and waypoint navigation. The main motivation behind this approach is a maximum reduction of the sim2real gap, avoiding the main pitfalls of end-to-end training of navigation in simulation followed by a transfer of neural models to the real environment. We explore an approach that prefers classical methods based on sensor models and optimization, motivated by their robustness, and employ machine learning in a targeted way for parts of the system where its use is both necessary and beneficial. We also limit input to trained models to representations with a potentially low sim2real gap. For this reason, during navigation the agent builds a metric bird's eye view occupancy map from the Lidar input and localizes itself on it using metric SLAM~\cite{TBF2002probabilisticrobotics}. This binary map is combined with an overlaid semantic point cloud, which contains the positions of key objects and their semantic classes, which are detected from the RGB input with an object detector. Detection and mapping are aligned through the SLAM algorithm's localization module.

\myParagraph{Navigation} is performed hierarchically on two different levels. On a higher level (outer
loop in Figure~\ref{fig:teaser}), 2D waypoint coordinates $\mathbf{p}_t{=}(x,y)$ are produced and provided to the lower level controller (inner loop), whose task is to navigate to the waypoint using the maintained occupancy map. The high-level controller switches between two different strategies:
\begin{itemize} 
\item[\ding{192}] \textbf{Exploration} --- when the target object has not yet been observed, i.e. the robot explores the environment, maximizing coverage. This is done with a learned policy trained with RL, see below.
\item[\ding{193}] \textbf{Exploitation} --- when the target object has been observed and thus is part of the semantic point cloud, its location is taken as a new waypoint and given to the local planner.
\end{itemize}



\myParagraph{Metric EgoMap} To gather navigability information along its path and more efficiently revisit previously seen areas, the agent builds what is called an EgoMap, an occupancy grid of fixed spatial resolution centered on its current position and aligned with its heading direction.

On the real robot, this map is obtained using the RTABMap~\cite{labbe19rtabmap} library. It uses a graph-based SLAM algorithm with loop closure, a flexible design taking advantage of RGB-D, Lidar and odometry sensor data. Lidar and/or depth are used to create a 2D/3D local occupancy grid, associated to a node whose initial position relies on odometry integration. Descriptors are then created from keypoints extracted from RGB frames in order to facilitate node comparison and loop closure detection. RTABMap also includes short- and long-term memory management, global map compression and multi-session mapping.

In simulation, we take advantage of privileged information to retrieve a complete top-down view of the scene navigability, through a projection of the NavMesh generated by the Recast\&Detour~\cite{recastdetour} library in Habitat-Sim. A fog-of-war mask is then built by ray-tracing in the agent's field of view directly on this top-down view using perfect localization.

Both real and sim approaches generate a global map on which we apply a simple affine transformation parameterized by the agent's current pose to get the EgoMap.

\myParagraph{Exploration} is the main module based on machine learning. In contrast to most recent work in embodied AI~\cite{BeechingECCV2020,Chaplot2020Learning}, the policy does not take the first person RGB input, but the EgoMap $\mathbf{M}_t$ produced by the metric SLAM algorithm. This leads to a significant simplification of the task and increased sample efficiency, and it minimizes the sim2real gap, as changes in lightning, color and texture are avoided. The policy is a part of the outer loop and predicts 2D waypoint coordinates $\mathbf{p}_t$. The problem is partially observable for multiple reasons: (i) not all areas of the scene have been observed at any point in time; (ii) for efficiency reasons, the EgoMap $\mathbf{M}_t$ does not cover the full scene, observed areas can therefore be forgotten when the agent navigates sufficiently far away from them; (iii) even theoretically fully observable problems (MDPs) can be transformed into POMDPs (``\textit{Epistemic POMDPs}'') in the presence of uncertainty in the environment, which is a standard case in robotics, as has recently been shown in~\cite{EpistemicPOMDPs2021}. We therefore imbued the policy with hidden memory $\mathbf{h}_t$ and made it recurrent.

\begin{figure}[t] \centering
\includegraphics[width=\linewidth]{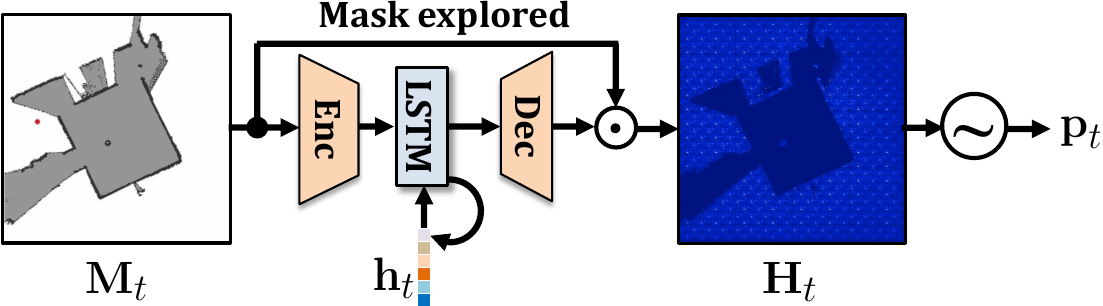}
\caption{\label{fig:exp_policy}The exploration policy takes as input EgoMaps $\mathbf{M}_t$ and predicts a heatmap, which is limited/masked ($\odot$) to unexplored areas. The next waypoint $\mathbf{p}_t$ is sampled ($\sim$) from the resulting map $\mathbf{H}_t$.}
\end{figure}

The policy $\pi$ needs to be able to predict multi-modal distributions, as there are multiple valid trajectories exploring an environment efficiently. We baked this into the policy through an inductive bias, which forces prediction to pass through a spatial heatmap $\mathbf{H}_t$, from which the chosen waypoint location is sampled. Before sampling, we restrict the heatmap to unexplored areas through masking. This choice also leads to a more interpretable model, as the distribution of targeted exploration points can be visualized (see Section~\ref{sec:experiments}). 
This can be formalized as follows (see also Figure \ref{fig:exp_policy}):
\begin{align}
    \mathbf{h}_t  &= \phi (\mathbf{M}_t, \mathbf{h}_{t-1}; \theta_{\phi})\\
    \mathbf{H}_t' &= \pi (\mathbf{h}_t; \theta_{\pi}), 
    \ \mathbf{H}_t = \mathbf{H}_t' \odot [ M_t==``\textrm{Unexplored}'']) \\
    \mathbf{p}_t &= \sim (\mathbf{H}_t),
\end{align}
where $\theta_{\pi}$ and $\theta_{\phi}$ are trainable parameters and $\phi$ is the update recurrent function of an LSTM with hidden state $\mathbf{h}_t$; gates have been omitted in the notation for simplicity. Here, $\textbf{p}_t{=}(x_t, y_t)$ is 2D coordinates of the point sampled in the spatial heatmap $\mathbf{H}_t$, `$\sim$' is the sampling operator. 

We train the exploration policy with RL to maximize coverage and use the following reward function $r_{t}$: 
\begin{equation}
    r_t = \alpha r_{de} , \quad r_{de} = e_t - e_{t-1}, 
\end{equation}
where $e_t$ denotes the explored area at step $t$,
$l$ is the number of inner environment steps necessary to navigate to the coordinates $(x,y)$ predicted by the policy, $\alpha$ is a scaling hyper-parameter set to $0.01$.

\myParagraph{Local navigation} to the waypoint $\mathbf{p}_t$ is performed by an analytical planner that computes the shortest path on the current occupancy EgoMap $\mathbf{M}_t$. This is not necessarily the optimal path, as the map is not equal to the (unobserved) GT map and the intermediate regions to be traversed (and even the waypoint $\mathbf{p}_t$) might be unexplored. We employ a dynamic planner D* which calculates the shortest path under classical assumptions and replans when new information is available. Since we optimize our method to be robust and efficient in real environments, unlike recent work \cite{ramakrishnan_poni_2022,Chaplot20objectgoal} we choose a D* planner over the commonly used Fast Marching Method~\cite{sethian1996fast}. The path feasibility in real and the speed of planning are the two main reasons behind this design choice.

\myParagraph{Stabilizing training} The potential failures and sub-optimal trajectories produced by local planning in uncertain conditions using D*, as described above, also negatively impact the training process of the exploration policy. This policy is a part of the outer loop and predicts waypoints $\mathbf{p}_t$, receiving a reward only upon completion of the full local navigation process. Noise in local planning impacts the stability of the RL training process and leads to lack of convergence. 

We solved this by training the exploration policy interfacing a local policy on which we imposed a length limit. The full trajectory from the current position to the next waypoint $\mathbf{p}_t$ predicted by the exploration policy is split into a sequence of small sub goals distanced by 0.3m, and the local policy is limited to 5 of these sub goals. Control is given back to the outer loop if the waypoint $\mathbf{p}_t$ has been reached, or the limit of 5 subgoals is reached. 
This choice lead to stable training and the trained policy transferred well to the targeted exploration task, without changes. The same limitation on the length of local planning is also applied at deployment, which led to improved robustness in real conditions and makes complex recovery behavior obsolete. 

\myParagraph{Object Detection and Mapping} is framed as a semantic segmentation task from the current RGB-D frame $o_{t}$, which we supervise from GT masks calculated from privileged information in the simulator. The predictor is a DeepLab v3 network~\cite{DeepLabv3}, detected objects in the mask are inversely projected and aligned with the EgoMap using depth information and the episodic odometry. Note that both, depth and odometry, are noisy in the real robot / real environment evaluation settings. 

\myParagraph{High-level decisions} are fully handcrafted, as this leads to a robust and transferable decision process where learning is arguably not required. Given our decision choices, only one type of decision is required, whether to perform exploration or exploitation (i.e. navigation towards the goal). This is taken on the basis whether the current target object has been observed at mapped, or not. If multiple objects of the same class have been detected, the location with the most probably detection (in terms of segmented object pixels) is chosen. A minimum number of pixels is required for an object to be mapped.

\begin{table}[t] \centering
{\small
\setlength\tabcolsep{2pt}
\begin{tabular}{a ccccc}
\toprule
Method & \multicolumn{2}{c}{--- LIDAR Usage ---}  & Aux & Obj.\\
       & Map & Low-lev cntrl & losses\cite{MarzaIROS2022} & Segm. \\
\midrule
ProjNMap+AUX \cite{MarzaIROS2022} & $-$& $\checkmark$ & $\checkmark$ & $-$\\
Ours & $\checkmark$ & $\checkmark$ & $-$ & $\checkmark$\\
\bottomrule
\end{tabular}
}
\caption{\label{tab:methods}Comparability of the different methods in terms of sensor and information availability. Both methods use LIDAR.}
\end{table}

\section{Experimental Results}
\label{sec:experiments}

\myParagraph{Simulation} for training in simulation and for additional evaluation (complementary to experiments on the real platform) we used the photo-realistic Habitat simulator~\cite{Savva_2019_ICCV} and two datasets with 3D scanned environments, each with the standard train/validation/test split: (i) the Gibson dataset (120 scenes) and the Matterport 3D dataset (90 scenes). 

\begin{figure}[t] \centering
\includegraphics[width=\linewidth]{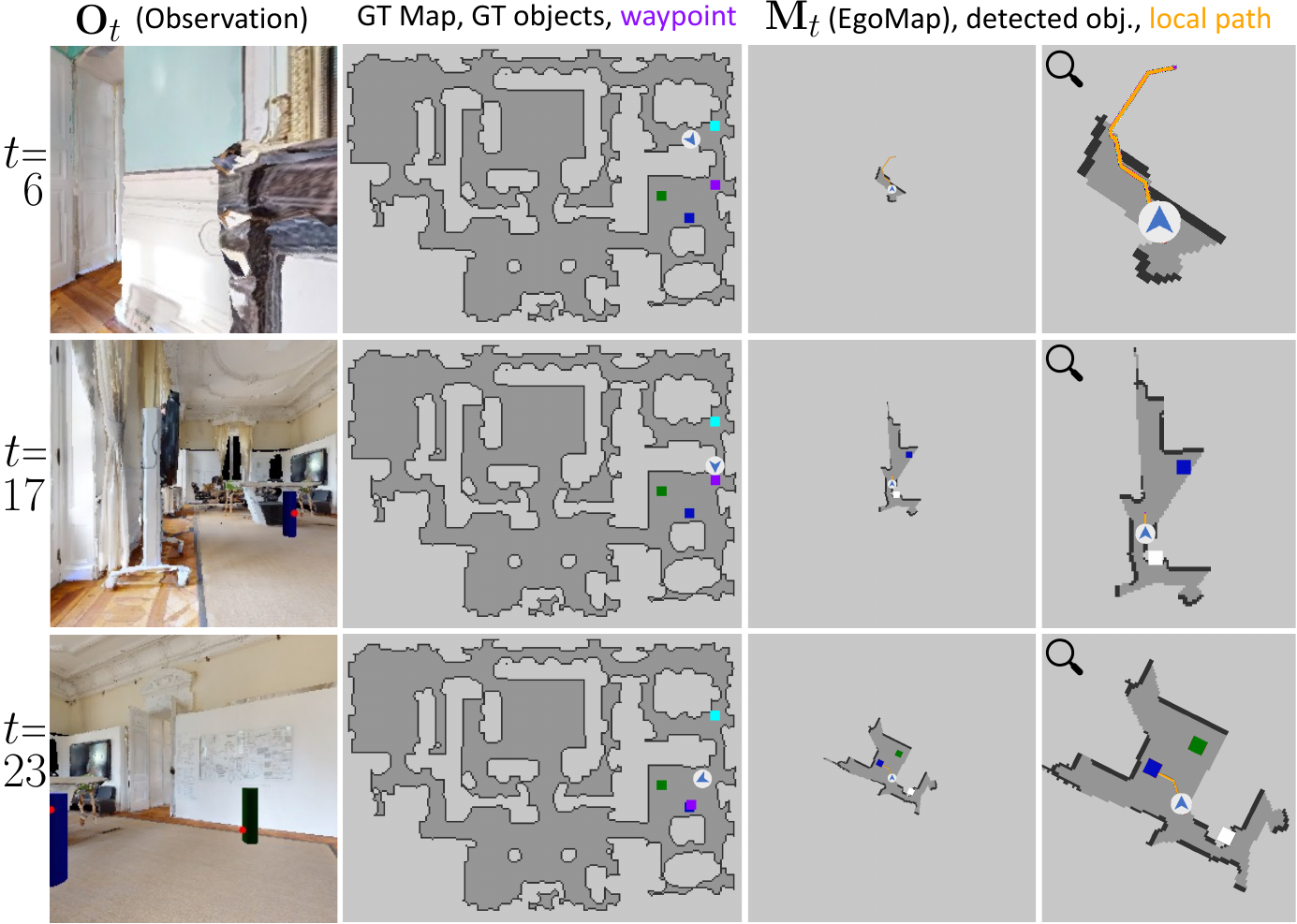}
\caption{\label{fig:qualitative_study} A rollout of an episode with the hybrid model. From left to right: (1) RGB observation; (2) GT map with the GT goal positions \includegraphics[height=0.8em]{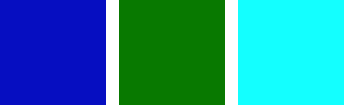}, the current agent position \includegraphics[height=0.8em]{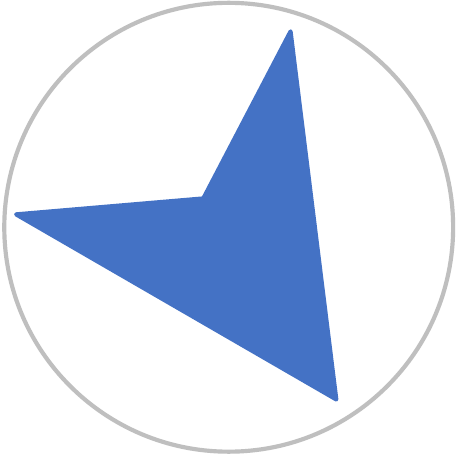}, the current \textcolor{way_point}{waypoint $p_t$} \includegraphics[height=0.8em]{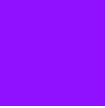}; (3) EgoMap $M_t$ with \textcolor{orange}{the planned local path} and (4) a zoomed version. The initial goal is blue. At $t{=}6$, an exploration goal is predicted. The agent enters a new room, and at $t{=}17$ it detects the \textcolor{blue_cylinder}{\textbf{blue goal}} and switches to exploitation mode advancing towards it. At $t{=}23$, it observes the very dark \textcolor{green_cylinder}{\textbf{green goal}} and maps it for future use. A false positive example (white cylinder) was also detected.}
\end{figure}

\myParagraph{Real environment} To evaluate the method in a real environment, we used a LoCoBot robot~\cite{locobot} [ \includegraphics[height=4mm]{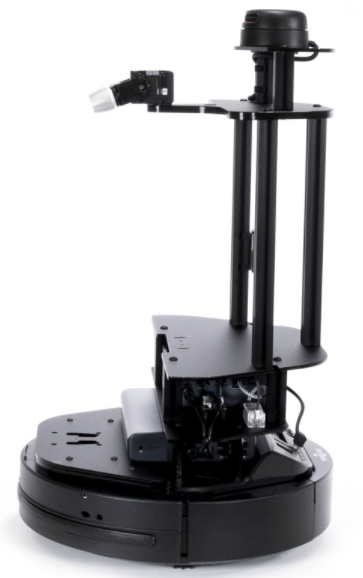} ] equipped with an Intel \textit{RealSense} RGB-D camera and a single-ray Lidar of type \textit{RPLIDAR A2M8} (see Figure \ref{fig:chateau}), which we restrict to Field-Of-View equivalent to the RGB camera. 
We used the publicly available \textit{habitat\_sim2real} library~\cite{HabitatSim2Real}, which allows to connect a real robot under \texttt{ROS} to the Habitat simulator as an agent.
We perform tests in a building with classical architecture covering one large conference room and several adjoining rooms, cf. Figure \ref{fig:chateau}, and for a map, Figure \ref{fig:qualitative_study}. The environment features difficult conditions including windows and glass panels, thick carpets, textureless walls, etc.

\myParagraph{Baselines} We compared with two end-to-end methods:
\begin{itemize}
    \item \textbf{ProjNMap} (Projective Neural Map) is based on projective neural memory as inductive bias for neural networks \cite{henriques18mapnet} and has been explored for the specific task of Multi-ON in \cite{DBLP:conf/nips/WaniPJCS20}, providing the best results in the original experiments when the task was introduced.
    \item \textbf{ProjNMap+AUX} combines the original ProjNMap model with auxiliary losses on an additional head, which predicts the direction and distance to the current target during training \cite{MarzaIROS2022}. This method achieved the winning performance in the CVPR 2021 M-ON Challenge~\cite{MONChallenge} and is the current state-of-the-art on this task.
\end{itemize}
Table \ref{tab:methods} summarizes sensor usage and information availability of the main baseline \cite{MarzaIROS2022} compared to our method. All methods use Lidar: ours - to maintain a metric map, the baseline \cite{MarzaIROS2022} - for a localization step necessary to perform closed-loop low-level control, mapping the discrete action space of the neural agent to the continuous motor space of the robot. All methods use information on object positions \textit{during training}: ours - in the pre-training step of the visual encoder, the baseline~\cite{MarzaIROS2022} - through auxiliary losses.

\myParagraph{Configurations}
With different evaluation goals in mind, we created two configurations of the agent:
\begin{itemize}
    \item \textbf{Locobot}: this configuration corresponds to the physical robot (Locobot) and its sensors. We also created a corresponding Habitat simulator configuration, which is equal to these settings: FOV of 56° (camera+Lidar), frame size of 160{$\times$}120
    and a compatible camera position. This configuration is of double use, i.e. can be used for evaluation in both simulation and the real environment.
    \item \textbf{Multi-ON-Compat}: we also reused the settings of the Multi-ON benchmark, making this configuration compatible with prior work like~\cite{MarzaIROS2022,DBLP:conf/nips/WaniPJCS20}. This includes a FOV of 79° and camera frames of size 256{$\times$}256. This configuration can be used in simulation only.
\end{itemize}

\begin{table*}[t] \centering
{\small
\begin{tabular}{a a  b b r r b b r r}
\toprule
\multicolumn{2}{c}{} & \multicolumn{4}{c}{------------ M-ON Setup ------------} & \multicolumn{4}{c}{------------ LoCo Setup ------------} \\
 Dataset & Agent &
 \textbf{Progress} & \textbf{PPL} & \textbf{Success} & \textbf{SPL} & 
 \textbf{Progress} & \textbf{PPL} & \textbf{Success} & \textbf{SPL}  \\
\hline
MP3D~ & ProjNMap  &
$41.63$ & 
$21.81$ & 
$23.10$ & 
$14.41$ &

$35.47$ & 
\textbf{23.97} & 
$19.10$ & 
$15.49$ 
\\
& ProjNMap+AUX \cite{MarzaIROS2022}  &
\textbf{62.47} & 
\textbf{35.21} & 
\textbf{48.20} & 
\textbf{62.47} &

$59.97$ & 
$23.29$ & 
$33.40$ & 
\textbf{19.75}
\\
& Hybrid (Ours) &
$55.23$ &
$12.41$ &
$41.80$ &
$11.72$ &

\textbf{63.37} & 
$19.72$ & 
\textbf{50.80} & 
$18.44$
\\
\hline
10 episodes & ProjNMap  &
$23.33$ & 
$14.39$ & 
$10.00$ & 
$9.21$ &

$16.67$ & 
\textbf{13.63} & 
$10.00$ & 
$9.52$
\\
& ProjNMap+AUX \cite{MarzaIROS2022}  &
$43.33$ & 
\textbf{28.19} & 
$30.00$ & 
\textbf{24.88} &

$40.00$ & 
$12.77$ & 
$30.00$ & 
$10.65$
\\
& Hybrid (Ours)  &
\textbf{50.00} &
$9.20$ &
\textbf{50.00} &
$9.20$ &

\textbf{56.67} & 
$12.09$ & 
\textbf{50.00} & 
\textbf{11.99}
\\
\hline
\midrule
\multicolumn{7}{l}{\footnotesize \textit{``M-ON'': CVPR 2021 Multi-On challenge sensor settings: FoV=79°, RGB size=$256{\times}256$}.} \\
\multicolumn{7}{l}{\footnotesize \textit{``LoCo'': Sensor settings equivalent to the physical robot: FoV=56°, RGB size=$160{\times}120$}.} \\
\bottomrule
\end{tabular}
}
\caption{\label{tab:multion_sim_results}Performance in Simulation (Habitat) for two different environments: the Matterport 3D validation set (comparable with the CVPR 2021 Multi-ON challenge), and 10 test episodes of the simulated version of our real environment.}
\vspace*{-5mm}
\end{table*}

\begin{table}[t] \centering
{\small
\begin{tabular}{ a b b c c}
\toprule
 Agent         &  \textbf{Progress} & \textbf{PPL} & \textbf{Success} & \textbf{SPL}  \\
\midrule
ProjNMap  & 
$3.00$ & 
$0.85$ & 
$0.00$ & 
$0.00$ 
\\
ProjNMap+AUX \cite{MarzaIROS2022}  & 
$0.00$ & 
$0.00$ & 
$0.00$ & 
$0.00$ 
\\
Hybrid (Ours) & 
$43.10$ & 
$6.02$ &
$20.00$ &
$4.99$ 
\\
\bottomrule
\end{tabular}
}
\caption{\label{tab:multion_real_results}Performance in the \textbf{real environment} by the physical Locobot on 10 test episodes. We compare  with the CVPR 2021 Multi-ON Challenge winner~\cite{MarzaIROS2022} (current SOTA).}
\vspace*{-5mm}
\end{table}

\begin{figure}[t] \centering
\includegraphics[width=0.9\linewidth]{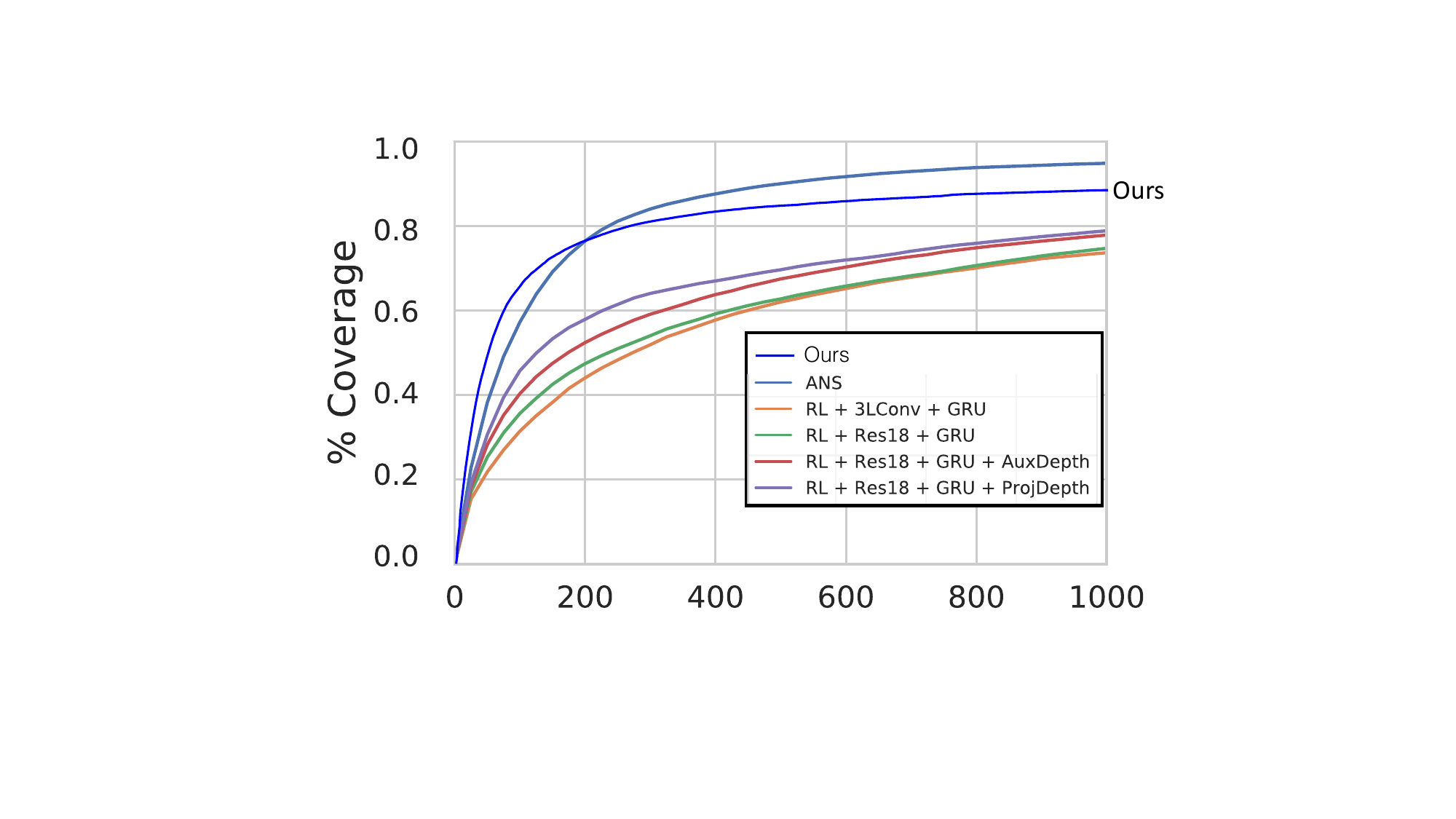}
\caption{\label{fig:coverageplot}Coverage (\%) obtained by the exploration policy as a function of episode length (the number of simulation steps), compared to ANS \cite{Chaplot2020Learning} and  end-to-end RL baselines using egocentric input taken from \cite{Chaplot2020Learning} on Gibson/Val.} 
\vspace*{-4mm}
\end{figure}

\begin{table}[t] \centering
{\small
\setlength\tabcolsep{2pt}
\begin{tabular}{a c|c}
\toprule
      & \multicolumn{2}{c}{--------- Coverage (\%) --------- } \\
Agent & Gibson & MP3D (domain generaliz.)\\
\midrule
RL+3LConv+GRU & 73.7 & 33.2\\
RL+Res18+GRU & 74.7 & 34.1 \\
RL+Res18+GRU+AuxDepth & 77.9 & 35.6 \\
RL+Res18+GRU+ProjDepth & 78.9 & 37.8 \\
ANS \cite{Chaplot2020Learning} & \textbf{94.8} & 52.1 \\
Ours & 88.4 & \textbf{67.14} \\
\bottomrule
\end{tabular}
}
\caption{\label{tab:exp_coverage}Coverage obtained by different exploration policies on Gibson and MP3D. All agents were trained on the Gibson train split (Results on competing methods taken from \cite{Chaplot2020Learning}).}
\vspace*{-5mm}
\end{table}

\myParagraph{Setup and hyper-parameters} 
decision thresholds are set as follows: at least 0.7\% of detected pixels is required for an object to be placed on a map; 5\% or more of detected pixels is required for an object to trigger the~\textit{Found} action.

\myParagraph{Results in Simulation} are shown in 
Table~\ref{tab:multion_sim_results}. They have been obtained on two different datasets: (i) on the validation split of the Matterport 3D dataset, making these runs comparable to the validation entries of the CVPR 2021 Multi-ON competition, and (ii) on 10 episodes in a 3D scanned version of our real environment, shown in Figure~\ref{fig:teaser}. These 10 episodes correspond to simulated versions of the episodes tested in the real environment, see further below. This simulated environment has \textit{not} been used for training.

While our main design choices are biased toward a robust performance in real environments, we can see that it is also highly performant in simulation. In the LoCo Setup, where the (virtual) sensor configuration mirrors real sensors, the method outperforms the state of the art on the \textit{Progress} and \textit{Success} metrics and is competitive in the others.

\myParagraph{Results in the real environment} are shown in Table~\ref{tab:multion_real_results}, on the same 10 episodes that we tested in simulation. Our hybrid agent was able to collect 43\% of the targets successfully and even finished 2 out of 10 episodes retrieving all 3 required items. We conjecture that this is due to the strategy to disentangle perception, exploration, and waypoint navigation, which allows for keeping the sim2real gap lower than what can be done for the end-to-end (E2E) trained methods.

On the other hand, the performance of the baselines can be considered a failure. While it has been reported, that end-to-end training of the simpler PointGoal task in similar conditions can be successful~\cite{SadekICRA2022}, this did not apply to our experiences on the much more complex Multi-ON task. While~\cite{MarzaIROS2022} obtained the state of the art in the official benchmark, i.e. in simulation, not a single episode was successful in the real environment. Most failure cases were related to the poor exploration of the scene and high uncertainty in detecting the targets. The agent had a hard time changing rooms and repeatedly failed to apply 'Found' when the object is closely upfront. We conjecture that the high impact of the sim2real gap on raw sensor data(cf Figure \ref{fig:chateau}) requires to adapt the E2E methods on real data for such complex tasks. Although training with real data is time-consuming, we believe that finetuning the E2E agents on offline real data with behavioral cloning is possible and would enhance the performance in upcoming real experiments.

\begin{figure}[t] \centering
\includegraphics[width=\linewidth]{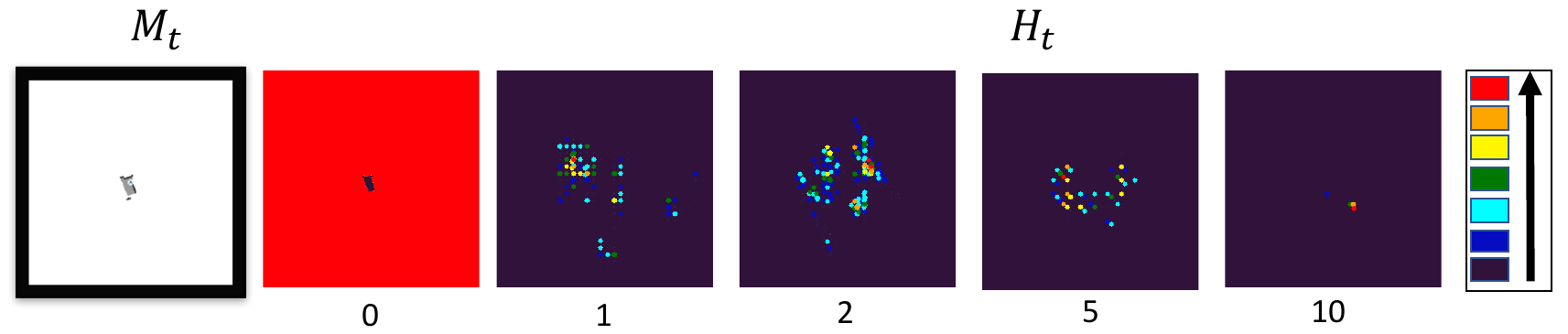}
\caption{\label{fig:qualitative_study_exp}For a given time step, we plot the predicted spatial heatmaps $\mathbf{H}_t$ for different training checkpoints, after 0,1,2,5 and 10 million updates. The lowest and highest probabilities are in dark blue and red, respectively.}
\end{figure}

\myParagraph{Exploration performance} is provided as complementary information in Figure~\ref{fig:coverageplot}. On this task, we compare with \textit{Active Neural SLAM} by Chaplot~\etal~\cite{Chaplot2020Learning}. 
We outperform it on the first 200 steps, making our method more robust for limited time-budget exploration. More importantly, given our main objective of optimizing sim2real performance, we outperform the state-of-the-art on domain generalization by +15\% margin, as shown in Table~\ref{tab:exp_coverage}. 

\myParagraph{Qualitative results} are given in Figure~\ref{fig:qualitative_study}, which shows a rollout for a single episode. We see that the agent first explores the scene, observes the goal quickly, and switches to exploitation mode. While navigating to the first goal, it observes a potential future goal and correctly maps it. Some drawbacks of the hybrid agent are presented in the detection of false positives such as the white cylinder.

In Figure~\ref{fig:qualitative_study_exp}, we visualize, for a chosen episode in a given environment step, the evolution of the predicted heatmaps $\mathbf{H}_t$ as training evolves (hence, different checkpoints). Heatmaps correspond to high-entropy distributions at the beginning, with high uncertainty on exploration targets. As training goes on, the distribution gets narrower and peaky, making the predictions more and more certain. 


\section{Conclusion}
\noindent 
 We have extended the Multi-ON task to real environments and up to our knowledge we present the first experimental evaluation of this task in these settings. We have introduced a hybrid model, which disentangles waypoint planning and semantics, significantly decreases the sim2real gap, and outperforms E2E trained models which were the current SOTA in simulation. Future work will focus on the enhancement of the handcrafted high-level strategy, which needs to be robust to the false positive detections, a common challenge also in other navigation tasks (e.g. ObjectNav).

\bibliographystyle{plain}
\bibliography{main}
\end{document}